\documentclass[lettersize,journal]{IEEEtran}
\usepackage{amsmath,amsfonts}
\usepackage{algorithmic}
\usepackage{booktabs}
\usepackage{multirow}
\usepackage{xcolor}
\usepackage{algorithm}
\usepackage{array}
\usepackage[caption=false,font=normalsize,labelfont=sf,textfont=sf]{subfig}
\usepackage{textcomp}
\usepackage{stfloats}
\usepackage{url}
\usepackage{verbatim}
\usepackage{graphicx}
\usepackage{cite}
\begin{document}

\title{Class-Adaptive Cooperative Perception for Multi-Class LiDAR-based 3D Object Detection in V2X Systems}

\author{Blessing Agyei Kyem, \and Joshua Kofi Asamoah, \and Armstrong Aboah\\
Department of Civil, Construction and Environmental Engineering\\
North Dakota State University\\
Fargo, ND 58102, USA}




\maketitle

\begin{abstract}
Cooperative perception enables connected vehicles and roadside infrastructure to share sensor observations, producing a fused scene representation that surpasses what any single platform can perceive alone. However, most cooperative 3D object detectors apply a uniform fusion strategy across all object categories, limiting their ability to accommodate the distinct geometric and sampling characteristics of different classes. This limitation is compounded by narrow evaluation protocols that typically focus on a single dominant class or a restricted set of cooperation patterns, leaving reliable multi-class detection across diverse vehicle-to-everything interaction modes insufficiently addressed. To bridge this gap, this paper presents a class-adaptive cooperative perception architecture for multi-class 3D object detection using LiDAR data. The architecture introduces four integrated components: multi-scale window attention with learned scale routing to capture features at varying spatial scales, a class-specific fusion stage that directs small and large objects through separate attentive pathways, bird's-eye-view enhancement through parallel dilated convolution and channel recalibration for multi-resolution context, and class-balanced objective weighting to counteract bias toward frequent categories. Extensive experiments on the V2X-Real benchmark span vehicle-centric, infrastructure-centric, vehicle-to-vehicle, infrastructure-to-infrastructure, and vehicle-to-infrastructure cooperation settings, with identical backbone and training configurations across all compared methods. The proposed system consistently outperforms strong intermediate-fusion baselines in mean detection quality, achieving the largest improvements on trucks, clear gains for pedestrians, and competitive performance on cars. These findings demonstrate that aligning fusion and feature extraction with class-dependent geometry and point density yields more balanced cooperative perception in realistic vehicle-to-everything deployments.
\end{abstract}


\section{Introduction}
\label{sec:introduction}
Autonomous driving has emerged as one of the most transformative technologies in intelligent transportation systems, with perception serving as the foundational module upon which planning and control decisions critically depend~\cite{han2024collaborative}. Despite significant advances in sensor technology and deep learning, single-agent perception remains fundamentally limited by occlusions, finite sensor range, and environmental clutter~\cite{xu2022v2xvit}. Cooperative perception has emerged as a promising paradigm to overcome these limitations, enabling connected vehicles and roadside infrastructure to share perceptual information and collectively construct a more complete understanding of the surrounding environment~\cite{han2024collaborative,xu2022opv2v}. The success of this paradigm hinges on how effectively multiple agents fuse and utilize one another's sensor data.

Vehicle-to-Everything (V2X) cooperative perception systems address this challenge by fusing sensor data from multiple agents, including vehicles and infrastructure, through strategies ranging from early feature-level fusion to late object-level fusion~\cite{xu2022v2xvit,chen2019fcooper}. Among these strategies, intermediate fusion has demonstrated strong detection performance~\cite{han2024collaborative}. Nevertheless, the field exhibits two important gaps. First, most existing work evaluates cooperative perception under a limited subset of collaboration modes, typically vehicle-to-vehicle (V2V) or vehicle-to-infrastructure (V2I), rather than across the full spectrum of real-world V2X scenarios, which encompasses vehicle-centric (VC), infrastructure-centric (IC), vehicle-to-vehicle (V2V), infrastructure-to-infrastructure (I2I), and vehicle-to-infrastructure (V2I) configurations~\cite{xiang2024v2xreal}. Compounding this limitation, cooperative 3D object detection has remained largely vehicle-centric: widely adopted benchmarks such as OPV2V~\cite{xu2022opv2v} and V2XSet~\cite{xu2022v2xvit} target vehicle detection exclusively, and the majority of methods~\cite{ma2023hydro3d,ren2024interruption,arnold2022coop3dod} are designed and evaluated with vehicles as the primary object class, leaving multi-class scenarios involving pedestrians, trucks, and other categories underexplored. Second, and more fundamentally, these methods apply a \emph{uniform} fusion strategy to all object classes, treating vehicles, pedestrians, and trucks identically despite their vastly different spatial scales, LiDAR return densities, and localization requirements~\cite{xu2022v2xvit,han2024collaborative}. This one-size-fits-all design may be suboptimal when object classes exhibit heterogeneous detection characteristics, and benchmark results lend support to this concern~\cite{oubouabdellah2025vru}.

Empirical evidence reveals pronounced performance disparities across classes \cite{lin2017focal, zhu2019cbgs, caesar2020nuscenes}. Vehicles, which occupy larger regions in bird's-eye view and receive abundant LiDAR returns, achieve relatively high detection accuracy \cite{lang2019pointpillars, yan2018second, yin2021centerpoint}. In contrast, pedestrians and trucks, representing smaller and larger scales respectively, often underperform \cite{peri2023longtailed}. These disparities arise from several factors: small objects are compressed into roughly a single feature cell by typical backbone downsampling \cite{lin2017fpn, zhou2018voxelnet}; sparse LiDAR coverage yields weaker feature responses for compact targets \cite{shi2020pvrcnn}; and vehicle-dominated datasets cause gradient signals from underrepresented classes to be overwhelmed during optimization. In light of such heterogeneity, the uniform-fusion assumption, that a single processing strategy suffices for all classes, appears inadequate. Yet to our knowledge, class-adaptive or class-specific fusion mechanisms for V2X cooperative perception remain largely unexplored~\cite{han2024collaborative}.

These problems motivate our central hypothesis: tailoring the fusion architecture and training objectives to the distinct requirements of different object classes can yield meaningful improvements in \emph{multi-class} detection performance. To test this hypothesis, we propose a \emph{class-adaptive cooperative perception architecture} that integrates four design principles into a unified framework. To align receptive fields with the spatial extent of each object class, the architecture employs multi-scale window attention. It further incorporates a class-specific fusion block that routes features through dedicated attention pathways tailored to small and large objects, a multi-scale bird's-eye view (BEV) enhancement module that captures context at multiple receptive scales, and a class-balanced loss that up-weights gradients from underrepresented classes during training~\cite{lin2017focal}. We evaluate this architecture on the V2X-Real dataset~\cite{xiang2024v2xreal}, the first large-scale real-world benchmark for V2X cooperative perception, across all five cooperation modes: VC, IC, V2V, I2I, and V2I. Under this evaluation protocol, the architecture achieves consistent improvements in mean average precision (mAP) across all cooperation types and object classes.

Our contributions are summarized as follows:

\begin{itemize}
    \item We propose a cooperative perception architecture that routes features through class-adaptive attention pathways, addressing the uniform-fusion assumption of existing V2X methods and improving multi-class detection performance.
    \item We integrate multi-scale window attention and multi-scale BEV enhancement, improving localization across object classes with disparate spatial scales.
    \item We adopt a class-balanced loss to counteract dataset imbalance, leading to gains for underrepresented classes while maintaining or improving overall mAP.
    \item We conduct comprehensive experiments on V2X-Real across all five cooperation types (VC, IC, V2V, I2I, V2I), providing detailed ablation studies and an analysis of class-specific performance and resolution limitations.
\end{itemize}

The remainder of this paper is organized as follows. Section~\ref{sec:related} reviews related work. Section~\ref{sec:method} presents our class-adaptive architecture. Section~\ref{sec:experiments} describes the experimental setup. Section~\ref{sec:results} provides comparative results, ablations, and analysis. Section~\ref{sec:conclusion} concludes the paper.

\section{Related Work}
\label{sec:related}
This section reviews prior work relevant to class-adaptive cooperative perception for multi-class 3D object detection. Section~\ref{subsec:coop_perception} surveys V2X cooperative perception methods and benchmarks. Section~\ref{subsec:multiclass_detection} examines multi-class 3D object detection challenges and solutions. Section~\ref{subsec:class_aware_arch} discusses class-adaptive and attention-based architectures.

\subsection{V2X Cooperative Perception}
\label{subsec:coop_perception}
Cooperative perception addresses the inherent limitations of single-agent systems (occlusions, limited sensing range, and sparse distant observations) by enabling connected vehicles and roadside infrastructure to share sensory information through Vehicle-to-Everything (V2X) communication~\cite{chen2019fcooper,xu2022opv2v,han2024collaborative}. The design space is shaped by fusion strategy: early fusion aggregates raw sensor data~\cite{arnold2022coop3dod}, late fusion merges independent detections~\cite{arnold2022coop3dod}, and intermediate fusion shares learned features~\cite{chen2019fcooper,han2024collaborative}, balancing perception accuracy with communication overhead. F-Cooper~\cite{chen2019fcooper} pioneered intermediate fusion with spatial attention for BEV feature aggregation, followed by V2VNet~\cite{wang2020v2vnet} for joint perception and prediction and When2com~\cite{liu2020when2com} for selective communication. Subsequent methods refined fusion mechanisms through learnable collaboration graphs in DiscoNet~\cite{li2021disconet}, spatial confidence maps in Where2comm~\cite{hu2022where2comm} for extreme bandwidth reduction, and transformer-based architectures in V2X-ViT~\cite{xu2022v2xvit} and CoBEVT~\cite{xu2022cobevt}, with HYDRO-3D~\cite{ma2023hydro3d} adding temporal reasoning via tracking history. Benchmarks have evolved from OPV2V's simulated V2V scenarios~\cite{xu2022opv2v} to real-world datasets including DAIR-V2X~\cite{yu2022dairv2x} for vehicle-infrastructure cooperation, V2V4Real~\cite{xu2023v2v4real} for multi-vehicle perception, and V2X-Real~\cite{xiang2024v2xreal}, which covers all five cooperation modes with multi-class annotations. Despite these advances, existing methods apply uniform fusion mechanisms to all object classes, whether through spatial attention~\cite{chen2019fcooper}, collaboration graphs~\cite{li2021disconet}, or transformers~\cite{xu2022v2xvit}, treating vehicles, pedestrians, and other categories identically despite their vastly different spatial scales, LiDAR return densities, and detection characteristics, motivating investigation into class-adaptive fusion mechanisms.

\subsection{Multi-Class 3D Object Detection}
\label{subsec:multiclass_detection}

Single-agent 3D object detection has progressed through successive innovations in point cloud representation and feature learning. Early voxel-based methods such as VoxelNet~\cite{zhou2018voxelnet} introduced end-to-end learning by encoding point clouds through voxel feature encoding layers, while SECOND~\cite{yan2018second} improved computational efficiency via sparse 3D convolutions. PointPillars~\cite{lang2019pointpillars} simplified this paradigm by encoding point clouds into vertical pillars and processing them with 2D convolutions, achieving real-time inference. In parallel, PointNet++~\cite{qi2017pointnetpp} demonstrated the value of hierarchical feature learning by recursively applying set abstraction on nested point cloud regions to capture multi-scale local structures. Methods integrating both representations emerged to combine their complementary strengths: PV-RCNN~\cite{shi2020pvrcnn} introduced point-voxel feature set abstraction to leverage efficient voxel convolutions alongside flexible point-based receptive fields, while Part-A2~\cite{shi2021parta2} exploited intra-object part supervision to improve fine-grained localization. Multi-scale feature aggregation became a recurring theme, with architectures borrowing from 2D detection advances such as Feature Pyramid Networks~\cite{lin2017fpn} to construct hierarchical representations that capture objects at varying scales. Despite these architectural refinements, multi-class 3D detection faces persistent challenges arising from heterogeneous object characteristics. Small objects such as pedestrians and cyclists suffer from sparse LiDAR returns and are easily compressed into single feature cells during backbone downsampling, while class imbalance in vehicle-dominated datasets causes gradient signals from underrepresented categories to be overwhelmed during optimization~\cite{lin2017focal,cui2019classbalanced}. Benchmark datasets such as nuScenes~\cite{caesar2020nuscenes} with 10 object categories and KITTI with vehicle, pedestrian, and cyclist annotations have exposed these performance disparities, revealing substantial gaps between vehicle detection accuracy and performance on vulnerable road users. To address these challenges, researchers have adopted solutions including focal loss~\cite{lin2017focal} to down-weight easy examples, class-balanced loss~\cite{cui2019classbalanced} to reweight gradients by the effective number of samples, and anchor-free detection frameworks such as FCOS~\cite{tian2019fcos}, CenterNet~\cite{zhou2019centernet}, and CenterPoint~\cite{yin2021centerpoint} that model objects as center points rather than predefined anchors, offering greater flexibility for diverse object scales. Attention-based architectures such as TANet~\cite{liu2020tanet} introduced triple attention mechanisms to enhance robustness for hard-to-detect pedestrians, while SA-SSD~\cite{he2020sassd} incorporated structure-aware auxiliary networks to preserve spatial information during downsampling. However, these techniques remain confined to single-agent detection pipelines and have not been adapted to cooperative perception fusion, where vehicles, pedestrians, and other categories continue to be processed through uniform feature aggregation mechanisms regardless of their distinct detection requirements.

\subsection{Class-Specific and Attention-Based Architectures}
\label{subsec:class_aware_arch}
The evolution of class-specific processing and attention mechanisms in object detection has yielded architectural patterns that could, in principle, inform cooperative perception design, yet remain unexplored in that context. Class-specific processing emerged prominently with Mask R-CNN~\cite{he2017maskrcnn}, which introduced separate mask prediction heads for each object category, enabling instance-level segmentation alongside detection by predicting class-specific masks through RoI-aligned features. This paradigm of per-class specialization was extended by Cascade R-CNN~\cite{cai2018cascadercnn}, which addressed quality mismatch in high-IoU detection through a sequence of detectors trained with progressively increasing IoU thresholds, and by Hybrid Task Cascade~\cite{chen2019htc}, which interweaved cascaded refinement across detection and segmentation tasks with direct mask feature connections between stages. In parallel, attention mechanisms gained prominence as a means to selectively emphasize informative features. Squeeze-and-Excitation networks~\cite{hu2018senet} pioneered channel-wise attention by modeling inter-channel dependencies through global pooling and gating operations, while CBAM~\cite{woo2018cbam} extended this with sequential channel and spatial attention modules to recalibrate feature responses across both dimensions. The advent of transformer architectures introduced self-attention as a fundamental building block for vision tasks: Vision Transformer~\cite{dosovitskiy2021vit} demonstrated that pure self-attention applied to image patches could rival convolutional networks when pre-trained at scale, while Swin Transformer~\cite{liu2021swin} introduced shifted window attention to achieve linear complexity by restricting self-attention to non-overlapping local windows with cross-window connections, enabling hierarchical feature learning at multiple scales. Deformable attention emerged as a computationally efficient alternative through Deformable DETR~\cite{zhu2021deformabledetr}, which attended to a sparse set of learnable sampling points rather than all feature locations, naturally supporting multi-scale feature aggregation. These attention mechanisms have been adopted in 3D detection, as evidenced by V2X-ViT's heterogeneous multi-agent attention~\cite{xu2022v2xvit} and TANet's triple attention~\cite{liu2020tanet}, yet they process all object classes uniformly without routing features through class-specific pathways. Multi-scale and multi-resolution architectures have similarly advanced feature representation: FPN~\cite{lin2017fpn} constructed top-down feature pyramids with lateral connections to build semantically strong representations at all scales, DeepLab~\cite{chen2017deeplab} employed atrous spatial pyramid pooling (ASPP) to capture multi-scale context through parallel dilated convolutions with varying rates, and BEVFormer~\cite{li2022bevformer} learned unified bird's-eye-view representations through spatiotemporal transformers that query multi-scale camera and LiDAR features. Flexible object representations such as RepPoints~\cite{yang2019reppoints}, which model objects as adaptive point sets rather than rigid boxes, further illustrate the potential for representation specialization. Despite this rich landscape of class-specific processing, attention mechanisms, and multi-scale architectures developed for single-agent detection, cooperative perception methods have not adopted these principles to differentiate fusion strategies across object classes with heterogeneous characteristics. This leaves a clear gap between the capabilities demonstrated in single-agent systems and the uniform-fusion paradigm that dominates cooperative perception research.

\section{Method}
\label{sec:method}

This section presents our class-adaptive cooperative perception architecture for multi-class 3D object detection in V2X systems. Section~\ref{subsec:formulation} formalizes the problem and provides an architecture overview. Sections~\ref{subsec:window_attention} through~\ref{subsec:loss} detail the four core components: multi-scale window attention, class-specific fusion block, multi-scale BEV enhancement, and class-balanced loss.

\subsection{Problem Formulation and Architecture Overview}
\label{subsec:formulation}

Consider a Vehicle-to-Everything (V2X) cooperative perception scenario comprising $N$ spatially distributed agents $\mathcal{A} = \{a_1, a_2, \ldots, a_N\}$, where each agent $a_i$ may be either an ego vehicle, a cooperative vehicle, or roadside infrastructure equipped with perception sensors. At time step $t$, each agent $a_i$ captures a local observation in the form of a LiDAR point cloud $\mathcal{P}_i \in \mathbb{R}^{M_i \times 4}$, where $M_i$ denotes the number of points and each point is represented by spatial coordinates $(x, y, z)$ and reflectance intensity. The agents share a common world coordinate frame established through localization and map-based alignment, enabling spatial correspondence of their observations despite heterogeneous viewpoints and sensor configurations. The objective of cooperative 3D object detection is to aggregate perceptual information from all agents to produce a unified set of 3D bounding boxes $\mathcal{B} = \{\mathbf{b}_j\}_{j=1}^{K}$ for the ego agent, where $K$ denotes the number of detected objects and each detection $\mathbf{b}_j = (\mathbf{c}_j, \mathbf{s}_j, \theta_j, \hat{c}_j, p_j)$ is parameterized by its center location $\mathbf{c}_j \in \mathbb{R}^3$, spatial extent $\mathbf{s}_j \in \mathbb{R}^3$, heading angle $\theta_j \in [0, 2\pi)$, predicted class label $\hat{c}_j \in \mathcal{C}$, and confidence score $p_j \in [0,1]$. The class space $\mathcal{C} = \{\text{pedestrian}, \text{car}, \text{truck}\}$ encompasses three object categories representing small vulnerable road users, standard-sized vehicles, and large vehicles respectively~\cite{xiang2024v2xreal}, each exhibiting distinct geometric scales and detection characteristics.

Existing cooperative perception methods adopt a uniform fusion paradigm where features from all agents are aggregated through a single, class-agnostic fusion function $\mathcal{F}_{\text{uni}}: \mathbb{R}^{N \times H \times W \times D} \rightarrow \mathbb{R}^{H \times W \times D}$, which processes bird's-eye-view (BEV) feature maps $\{\mathbf{F}_i \in \mathbb{R}^{H \times W \times D}\}_{i=1}^{N}$ identically regardless of the object classes they represent. Here, $H$, $W$, and $D$ denote the spatial height, width, and channel dimensions of the BEV feature space, respectively. This uniform treatment implicitly assumes that the optimal fusion strategy is invariant across object categories, despite empirical evidence that vehicles, pedestrians, and other classes require fundamentally different receptive fields and feature resolutions due to their heterogeneous spatial scales and LiDAR return densities. Small objects such as pedestrians, which occupy limited spatial extent and yield sparse point observations, benefit from fine-grained feature representations and localized attention, while large objects such as vehicles and trucks, with abundant geometric structure, require broader contextual aggregation. The uniform fusion assumption fails to account for these class-specific requirements, resulting in suboptimal detection performance across the multi-class spectrum.

\begin{figure*}[!t]
\centering
\includegraphics[width=0.95\textwidth]{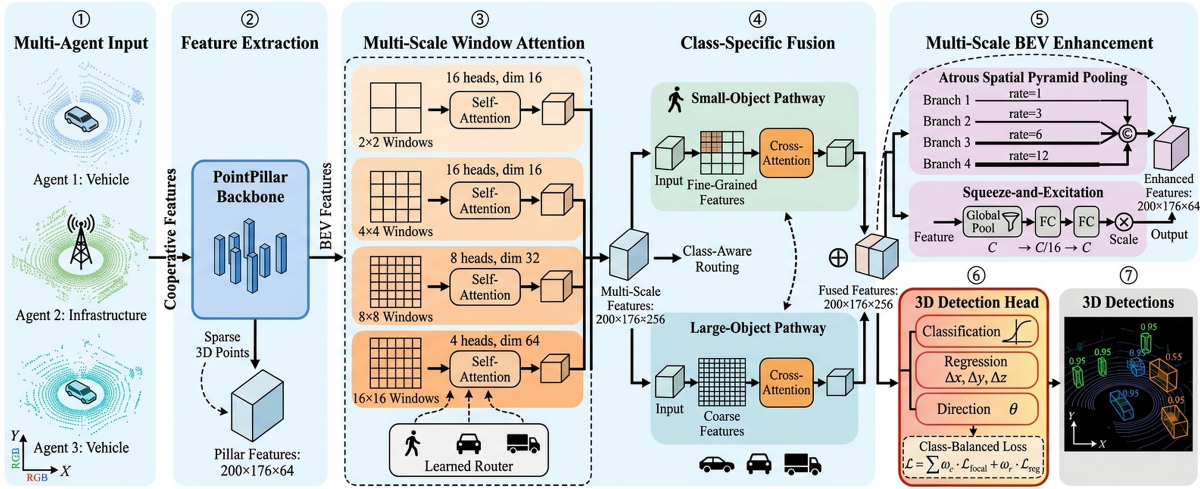}
\caption{Overview of the proposed class-adaptive cooperative perception architecture for multi-class 3D object detection in V2X systems. The framework processes multi-agent LiDAR inputs through: (1) Multi-agent feature extraction via PointPillars backbone, (2) Multi-scale window attention with learned class-adaptive routing that adapts receptive fields to object scales (2×2, 4×4, 8×8, 16×16 windows), (3) Class-specific fusion block with dedicated pathways for small objects (pedestrians) and large objects (cars, trucks), (4) Multi-scale BEV enhancement using atrous spatial pyramid pooling (ASPP) with dilation rates [1, 3, 6, 12] and squeeze-and-excitation (SE) modules, and (5) 3D detection head with class-balanced loss optimization. The architecture addresses scale variation and class imbalance through differentiated feature routing and gradient reweighting.}
\label{fig:architecture}
\end{figure*}

Formally, we replace the uniform fusion function $\mathcal{F}_{\text{uni}}$ with a class-adaptive fusion operator $\mathcal{F}_{\text{class}}$ that conditions feature aggregation on semantic class information:
\begin{equation}
\mathbf{F}_{\text{fused}} = \mathcal{F}_{\text{class}}\left(\{\mathbf{F}_i\}_{i=1}^{N} \mid \mathcal{C}\right),
\end{equation}
where $\mathbf{F}_{\text{fused}}$ denotes the fused feature representation and $\mathcal{C}$ provides class-specific routing signals derived from preliminary object classification. The fused features are subsequently processed by a detection head to produce the final set of 3D bounding boxes $\mathcal{B}$. This class-differentiated paradigm enables the architecture to allocate computational and representational capacity according to the distinct requirements of each object category, improving detection accuracy across the full multi-class spectrum while maintaining computational efficiency through selective feature routing. The following subsections detail each architectural component and describe the training procedure that optimizes the class-adaptive fusion mechanism end-to-end.

\subsection{Multi-Scale Window Attention}
\label{subsec:window_attention}

Objects of different categories exhibit substantial variation in spatial extent, with pedestrians occupying approximately $0.6 \times 0.6$ meters in BEV space while vehicles span $4 \times 2$ meters or more. Standard self-attention mechanisms~\cite{dosovitskiy2021vit} with fixed receptive fields fail to accommodate this heterogeneity, applying uniform window sizes that are either too coarse for small objects or computationally prohibitive for large ones. To address this, we introduce a multi-scale window attention mechanism that adapts the receptive field according to object scale, enabling efficient feature aggregation tailored to each category's geometric characteristics.

Given a BEV feature map $\mathbf{F} \in \mathbb{R}^{H \times W \times D}$ from agent $a_i$, we employ a pyramid of window sizes $\mathcal{W} = \{w_1, w_2, w_3, w_4\} = \{2, 4, 8, 16\}$ grid cells to capture features at multiple spatial scales. For each window size $w_s \in \mathcal{W}$, the feature map is partitioned into $\frac{H}{w_s} \times \frac{W}{w_s}$ non-overlapping windows, and self-attention is computed independently within each window $\mathcal{W}_k$:
\begin{equation}
\text{Attention}(\mathbf{Q}_k, \mathbf{K}_k, \mathbf{V}_k) = \text{softmax}\left(\frac{\mathbf{Q}_k \mathbf{K}_k^\top}{\sqrt{d}}\right) \mathbf{V}_k,
\end{equation}
where $\mathbf{Q}_k, \mathbf{K}_k, \mathbf{V}_k \in \mathbb{R}^{w_s^2 \times d}$ are the query, key, and value matrices for window $\mathcal{W}_k$ at scale $s$, obtained through learned linear projections with attention head dimension $d$. Each scale employs a pyramid of attention heads: fine scales ($w_1=2, w_2=4$) use 16 heads with dimension 16, while coarse scales ($w_3=8, w_4=16$) use progressively fewer heads (8, 4) with larger dimensions (32, 64) to balance receptive field coverage with computational cost. The window-based computation reduces complexity from $O(H^2W^2)$ for global attention to $O(w_s^2 \cdot HW)$ per scale, maintaining efficiency while capturing multi-scale context.

To enable interaction across windows, we adopt shifted window attention~\cite{liu2021swin} where alternating layers displace windows by $\lfloor w_s / 2 \rfloor$ grid cells, creating overlapping receptive fields. The multi-scale outputs $\{\mathbf{F}^{(s)}\}_{s=1}^{4}$ are adaptively fused through class-adaptive routing: a learned router network predicts class-conditional scale weights $\boldsymbol{\omega}(\mathbf{p}) \in \mathbb{R}^{4}$ at each spatial location $\mathbf{p}$, enabling the architecture to emphasize fine scales in regions containing pedestrians and coarse scales for vehicles and trucks. The routed output is computed as $\mathbf{F}_{\text{routed}} = \sum_{s=1}^{4} \omega_s(\mathbf{p}) \mathbf{F}^{(s)}(\mathbf{p})$, producing scale-adapted features that form the input to the subsequent class-specific fusion block.

\subsection{Class-Specific Fusion Block}
\label{subsec:class_fusion}

The class-specific fusion block operationalizes the core principle of differentiated feature aggregation by routing features from cooperating agents through dedicated processing pathways tailored to object characteristics. Rather than applying a single fusion operation uniformly across all categories, we partition the class space into two groups based on geometric scale: small objects $\mathcal{C}_{\text{small}} = \{\text{pedestrian}\}$ and large objects $\mathcal{C}_{\text{large}} = \{\text{car}, \text{truck}\}$. Pedestrians, representing vulnerable road users with compact spatial footprints (typically $< 1$ m$^2$ in BEV), require fine-grained localization and detailed feature preservation, while cars and trucks, spanning larger spatial extents ($> 4$ m$^2$), benefit from broader contextual aggregation. Each group is processed through a specialized attention branch that adapts the fusion mechanism to the spatial and structural properties of its constituent classes.

For the ego agent $a_{\text{ego}}$ seeking to fuse information from $N-1$ cooperative agents, we denote the class-specific feature map from agent $a_i$ for object group $g \in \{\text{small}, \text{large}\}$ as $\mathbf{F}_i^g \in \mathbb{R}^{H \times W \times D}$. The fusion process employs cross-attention~\cite{xu2022v2xvit} to aggregate features from cooperating agents based on spatial and semantic relevance. For each group $g$, the ego agent's feature map $\mathbf{F}_{\text{ego}}^g$ serves as the query, while features from all agents $\{\mathbf{F}_i^g\}_{i=1}^{N}$ provide keys and values. The cross-attention operation is formulated as:
\begin{equation}
\mathbf{F}_{\text{fused}}^g = \text{CrossAttn}\left(\mathbf{Q}_{\text{ego}}^g, \{\mathbf{K}_i^g, \mathbf{V}_i^g\}_{i=1}^{N}\right),
\end{equation}
where $\mathbf{Q}_{\text{ego}}^g = \mathbf{W}_Q^g \mathbf{F}_{\text{ego}}^g$, $\mathbf{K}_i^g = \mathbf{W}_K^g \mathbf{F}_i^g$, and $\mathbf{V}_i^g = \mathbf{W}_V^g \mathbf{F}_i^g$ are obtained through learned projection matrices $\mathbf{W}_Q^g, \mathbf{W}_K^g, \mathbf{W}_V^g \in \mathbb{R}^{D \times d}$ specific to group $g$. The cross-attention weights are computed as:
\begin{equation}
\alpha_{ij}^g = \frac{\exp\left(\frac{\mathbf{Q}_{\text{ego}}^g(\mathbf{p}) \cdot \mathbf{K}_i^g(\mathbf{p})}{\sqrt{d}}\right)}{\sum_{k=1}^{N} \exp\left(\frac{\mathbf{Q}_{\text{ego}}^g(\mathbf{p}) \cdot \mathbf{K}_k^g(\mathbf{p})}{\sqrt{d}}\right)},
\end{equation}
where $\mathbf{p} = (h, w)$ indexes spatial locations in the BEV grid with $h \in \{1, \ldots, H\}$ and $w \in \{1, \ldots, W\}$. The attention weights $\alpha_{ij}^g$ reflect the relevance of agent $a_i$'s observation to the ego agent's perception of objects in group $g$ at location $\mathbf{p}$, enabling adaptive information selection based on viewpoint complementarity and occlusion geometry.

The key distinction between small-object and large-object pathways lies in their architectural parameterization. Small-object fusion employs higher-capacity attention heads and finer spatial resolution to preserve detailed localization cues critical for pedestrian detection, while large-object fusion uses wider receptive fields to aggregate broader contextual information suitable for vehicle localization. Following group-specific fusion, the features are concatenated channel-wise and refined through a shared convolutional layer to produce the unified cooperative feature representation:
\begin{equation}
\mathbf{F}_{\text{coop}} = \text{Conv}\left(\text{Concat}\left(\mathbf{F}_{\text{fused}}^{\text{small}}, \mathbf{F}_{\text{fused}}^{\text{large}}\right)\right),
\end{equation}
where $\mathbf{F}_{\text{coop}} \in \mathbb{R}^{H \times W \times D}$ integrates class-differentiated fusion outputs into a cohesive representation. This design ensures that small and large objects benefit from fusion strategies matched to their detection requirements, while maintaining a unified feature space for downstream processing. The class-specific routing mechanism thus addresses the uniform-fusion limitation by explicitly conditioning aggregation operations on object scale and structural characteristics.

\subsection{Multi-Scale BEV Enhancement}
\label{subsec:bev_enhancement}

Accurate multi-class detection requires capturing contextual information at multiple spatial scales to accommodate the diverse geometric extents of pedestrians, cars, and trucks. While the class-specific fusion block provides differentiated feature aggregation, the fused representation $\mathbf{F}_{\text{coop}}$ may still lack sufficient multi-resolution context for robust localization across the full object size spectrum. To address this, we introduce a multi-scale BEV enhancement module that enriches the cooperative features through parallel multi-scale context aggregation and channel-wise recalibration.

The enhancement module employs an Atrous Spatial Pyramid Pooling (ASPP) structure~\cite{chen2017deeplab} to capture multi-scale contextual information through parallel atrous convolutions with varying dilation rates. Given the cooperative feature map $\mathbf{F}_{\text{coop}} \in \mathbb{R}^{H \times W \times D}$, we apply $L=4$ parallel convolutional branches with dilation rates $\{r_1, r_2, r_3, r_4\} = \{1, 3, 6, 12\}$ to extract features at receptive field scales ranging from 0.4 meters (single-cell, pedestrian-scale) to 9.6 meters (truck-scale):
\begin{equation}
\mathbf{F}_{\text{ASPP}}^{(l)} = \text{Conv}_{r_l}(\mathbf{F}_{\text{coop}}), \quad l \in \{1, \ldots, 4\},
\end{equation}
where $\text{Conv}_{r_l}$ denotes a $3 \times 3$ atrous convolution with dilation rate $r_l$ and $\mathbf{F}_{\text{ASPP}}^{(l)} \in \mathbb{R}^{H \times W \times D'}$ represents the feature response at scale $l$, with $D'=64$ denoting the output channel dimension per branch. An additional global average pooling branch captures scene-level contextual features. The multi-scale features are concatenated along the channel dimension:
\begin{equation}
\mathbf{F}_{\text{concat}} = \text{Concat}\left(\mathbf{F}_{\text{ASPP}}^{(1)}, \ldots, \mathbf{F}_{\text{ASPP}}^{(L)}, \mathbf{F}_{\text{global}}\right),
\end{equation}
where $\mathbf{F}_{\text{global}} \in \mathbb{R}^{H \times W \times D'}$ represents the upsampled global pooling features and $\mathbf{F}_{\text{concat}} \in \mathbb{R}^{H \times W \times 5D'}$ aggregates context across five parallel branches. The concatenated features are projected back to the original dimension through a $1 \times 1$ convolution with batch normalization and ReLU activation.

To adaptively recalibrate channel responses and emphasize informative features for multi-class detection, we integrate a Squeeze-and-Excitation (SE) block~\cite{hu2018senet} following the multi-scale aggregation. The SE module computes channel-wise attention weights through global average pooling followed by a two-layer fully connected network:
\begin{equation}
\mathbf{s} = \sigma\left(\mathbf{W}_2 \delta\left(\mathbf{W}_1 \mathbf{z}\right)\right),
\end{equation}
where $\mathbf{z} = \frac{1}{HW} \sum_{h=1}^{H} \sum_{w=1}^{W} \mathbf{F}_{\text{proj}}(h, w, :) \in \mathbb{R}^{D}$ encodes global channel statistics from the projected ASPP features $\mathbf{F}_{\text{proj}}$, $\mathbf{W}_1 \in \mathbb{R}^{D/r \times D}$ and $\mathbf{W}_2 \in \mathbb{R}^{D \times D/r}$ are learned projection matrices with reduction ratio $r=16$, $\delta(\cdot)$ denotes ReLU activation, $\sigma(\cdot)$ is the sigmoid function, and $\mathbf{s} \in \mathbb{R}^{D}$ represents the channel-wise scaling factors. The final enhanced representation incorporates a residual connection:
\begin{equation}
\mathbf{F}_{\text{BEV}} = \mathbf{F}_{\text{coop}} + \mathbf{s} \odot \mathbf{F}_{\text{proj}},
\end{equation}
where $\odot$ denotes element-wise multiplication broadcast across spatial dimensions. The residual connection preserves the original cooperative features while augmenting them with multi-scale context and channel recalibration, ensuring stable training and improved localization across the full range of object scales.

\subsection{Class-Balanced Loss}
\label{subsec:loss}

Multi-class 3D detection suffers from severe class imbalance, where vehicle-dominated datasets cause gradient signals from underrepresented categories to be overwhelmed during optimization, resulting in poor detection performance for pedestrians and trucks. To address this, we adopt a class-balanced loss that reweights the contribution of each class based on its representation in the training set, ensuring balanced gradient flow across all categories.

The total loss combines focal loss~\cite{lin2017focal} for classification, weighted smooth L1 loss for bounding box regression, and directional classification loss, each reweighted by class-specific factors:
\begin{equation}
\mathcal{L}_{\text{total}} = \sum_{c \in \mathcal{C}} \left( \omega_c^{\text{cls}} \mathcal{L}_{\text{focal}}^{(c)} + \omega_c^{\text{reg}} \mathcal{L}_{\text{reg}}^{(c)} + \omega_c^{\text{dir}} \mathcal{L}_{\text{dir}}^{(c)} \right),
\end{equation}
where $\omega_c^{\text{cls}}$ and $\omega_c^{\text{reg}}$ are class-specific weights inspired by the effective number of samples~\cite{cui2019classbalanced}. The effective number formulation computes weights as $\omega_c \propto (1 - \beta)/(1 - \beta^{n_c})$, where $n_c$ is the sample count for class $c$ and $\beta \in [0,1)$ controls the degree of reweighting. As $\beta \to 1$, the weighting approaches inverse class frequency, while $\beta \to 0$ reduces to uniform weighting. For the V2X-Real dataset, we set $\boldsymbol{\omega}^{\text{cls}} = [1.0, 3.0, 1.5]$ and $\boldsymbol{\omega}^{\text{reg}} = [1.0, 2.0, 1.0]$ for car, pedestrian, and truck classes respectively, giving pedestrians three times the classification weight and twice the regression weight relative to cars. This reweighting counteracts dataset imbalance by amplifying gradient signals from underrepresented classes during backpropagation, improving detection performance for pedestrians and trucks without sacrificing accuracy on the dominant car category.

\section{Experiments}
\label{sec:experiments}

This section presents comprehensive experimental validation of the proposed class-adaptive cooperative perception architecture. Section~\ref{subsec:exp_setup} describes the dataset, evaluation metrics, and implementation details. Section~\ref{subsec:comparative} presents comparative results across all cooperation modes. Section~\ref{subsec:qualitative} provides qualitative analysis. Section~\ref{subsec:cost} reports computational cost. Section~\ref{subsec:ablation} validates each architectural component through ablation studies. Section~\ref{subsec:range} provides range-stratified analysis.

\subsection{Experimental Setup}
\label{subsec:exp_setup}

\textbf{Dataset and Splits.} We evaluate our method on V2X-Real~\cite{xiang2024v2xreal}, the first large-scale real-world dataset for V2X cooperative perception. The dataset comprises 33,000 LiDAR frames with 1.2 million annotated 3D bounding boxes across 10 fine-grained categories, grouped into three super-classes: pedestrian, car, and truck. Following the official protocol, we use the train/validation/test splits containing 23,379/2,770/6,850 frames respectively. All models are trained once on the full training set and evaluated across five cooperation mode variants during inference: vehicle-centric, infrastructure-centric, vehicle-to-vehicle, infrastructure-to-infrastructure, vehicle-to-infrastructure. Each mode involves up to $N=4$ cooperating agents including two vehicles and two infrastructure units, enabling systematic evaluation of class-adaptive fusion under diverse V2X configurations.

\begin{figure}[t]
\centering
\includegraphics[width=\columnwidth]{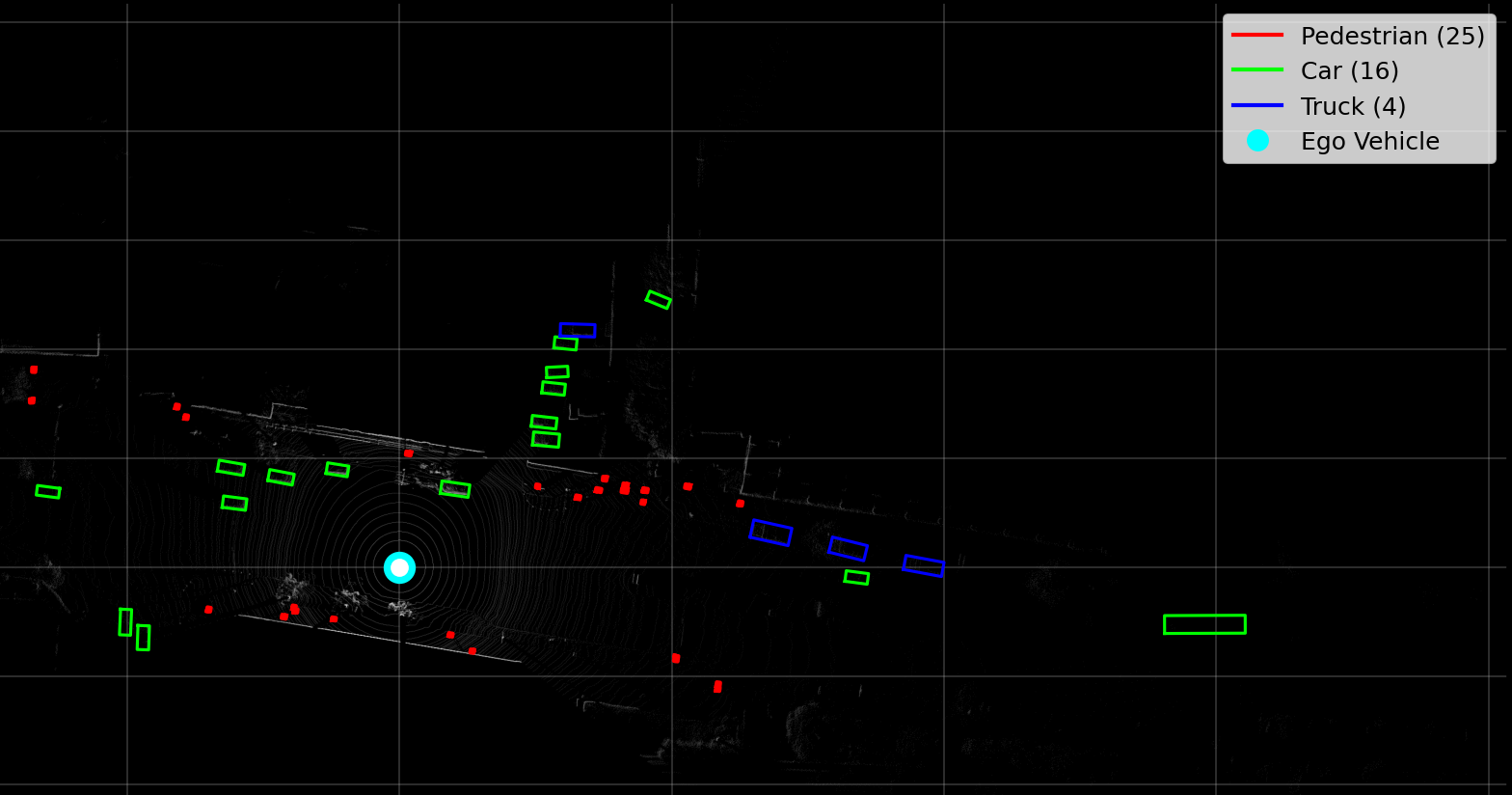}
\caption{Bird's-eye-view visualization of a representative V2X-Real scene showing multi-class 3D annotations: pedestrians (red, 25), cars (green, 16), and trucks (blue, 4) relative to ego vehicle (cyan). The dataset exhibits significant class imbalance and scale variations. }
\label{fig:sample_bev_image}
\end{figure}

\textbf{Baseline Methods.} We compare our approach against three intermediate fusion baselines, all adapted to support multi-class detection on V2X-Real with identical backbone configurations and training procedures: \textbf{F-Cooper}~\cite{chen2019fcooper}, \textbf{AttFuse}~\cite{xu2022opv2v}, and \textbf{V2X-ViT}~\cite{xu2022v2xvit}. All baselines use standard focal loss without class-specific reweighting, providing a controlled comparison to isolate the benefits of our class-adaptive design.

\textbf{Evaluation Metrics.} Detection performance is evaluated within a spatial region of $[-100, 100]$ meters in the $x$-direction and $[-40, 40]$ meters in the $y$-direction relative to the ego agent's coordinate frame. We compute Average Precision (AP) for each class $c \in \mathcal{C}$ at Intersection-over-Union (IoU) thresholds of 0.3 and 0.5, following the V2X-Real protocol~\cite{xiang2024v2xreal}. These lower thresholds are adopted to account for the substantial size variation across categories, spanning from sub-meter pedestrians to large trucks exceeding 10 meters. The mean Average Precision (mAP) is computed as the unweighted average across the three classes: $\text{mAP} = \frac{1}{3}(\text{AP}_{\text{ped}} + \text{AP}_{\text{car}} + \text{AP}_{\text{truck}})$. We report both AP@0.3 and AP@0.5 to evaluate detection quality at different localization precision levels.

\textbf{Implementation Details.} We adopt PointPillars~\cite{lang2019pointpillars} as the LiDAR backbone with a voxel size of 0.4 meters in both horizontal directions and 4 meters vertically. Point clouds are encoded into vertical pillars with a maximum of 32 points per pillar and 64,000 pillars per frame during training (70,000 during testing), producing BEV feature maps with spatial dimensions $H \times W = 352 \times 96$ grid cells covering $[-140.8, 140.8]$ meters in the $x$-direction and $[-38.4, 38.4]$ meters in the $y$-direction. After backbone processing with layer strides [2, 2, 2], the final BEV features have dimensions $H \times W = 200 \times 48$ with channel dimension $D=256$. The cooperative perception framework is implemented using OpenCOOD~\cite{xu2022opv2v}, extended to support multi-class detection through separate anchor configurations and classification heads for each category. For the multi-scale window attention mechanism, we employ a pyramid of four window sizes $\mathcal{W} = \{2, 4, 8, 16\}$ grid cells with attention heads $[16, 16, 8, 4]$ and head dimensions $[16, 16, 32, 64]$ respectively. Class-adaptive routing dynamically weights these scales based on predicted object classes at each spatial location. The ASPP module employs $L=4$ parallel branches with dilation rates $\{1, 3, 6, 12\}$ and output channel dimension $D'=64$ per branch. The SE block uses reduction ratio $r=16$. For class-balanced loss, we set classification weights $\boldsymbol{\omega}^{\text{cls}} = [1.0, 3.0, 1.5]$ and regression weights $\boldsymbol{\omega}^{\text{reg}} = [1.0, 2.0, 1.0]$ for car, pedestrian, and truck classes respectively.

\textbf{Training Procedure.} All models are trained for 80 epochs using the Adam optimizer~\cite{kingma2015adam} with an initial learning rate of $\eta_0 = 0.001$, weight decay $\lambda_w = 10^{-4}$, epsilon $\epsilon = 10^{-10}$, and batch size of 2. A multi-step learning rate scheduler reduces the learning rate by a factor of $\gamma=0.1$ at milestones [10, 50] epochs. Data augmentation strategies include random horizontal flipping, random rotation within $[-\pi/4, \pi/4]$ radians, and random scaling with factors sampled from $[0.95, 1.05]$, applied consistently across all agents to preserve spatial correspondence. The loss function combines focal loss ($\alpha=0.25$, $\gamma=2.0$) for classification, weighted smooth L1 loss ($\sigma=3.0$) for regression, and softmax classification loss for heading direction, with relative loss weights [1.0, 2.0, 0.2] respectively. Training is performed on NVIDIA H100 GPUs with mixed-precision enabled.

\textbf{Inference Pipeline.} During inference, each agent extracts BEV features from its local point cloud using the PointPillars backbone. Features are projected to the ego agent's coordinate frame via pose transformation matrices obtained from GPS/IMU localization. The ego agent applies multi-scale window attention with class-adaptive routing to each agent's features, aggregates them through class-specific fusion pathways, and enhances the fused representation via the ASPP-SE module with residual connection. The detection head produces class-specific predictions using anchor-based regression, and per-class non-maximum suppression (NMS) with IoU threshold 0.15 generates the final detection set $\mathcal{B}$.

\section{Results and Discussion}
\label{sec:results}

This section presents a comprehensive evaluation of the proposed class-adaptive cooperative perception architecture. We begin with comparative results against state-of-the-art intermediate fusion baselines in Section~\ref{subsec:comparative}, followed by qualitative results in Section~\ref{subsec:qualitative}. Section~\ref{subsec:cost} reports computational cost, Section~\ref{subsec:ablation} provides ablation studies, and Section~\ref{subsec:range} presents range-stratified analysis.

\subsection{Comparative Results}
\label{subsec:comparative}

Table~\ref{tab:main_results} presents the detection performance of our method compared to three state-of-the-art intermediate fusion baselines: F-Cooper~\cite{chen2019fcooper}, AttFuse~\cite{xu2022opv2v}, and V2X-ViT~\cite{xu2022v2xvit}. All methods are evaluated across five cooperation modes on the V2X-Real test set, with results reported as Average Precision at IoU thresholds of 0.3 and 0.5 for each class. Our analysis reveals three key findings regarding the effectiveness of class-adaptive cooperative perception.

\textbf{Consistent improvements across all cooperation modes.} Our method achieves superior mAP@0.5 across all five cooperation modes, with gains from +1.5 points (I2I: 43.9\% vs. 42.4\% V2X-ViT) to +6.2 points (V2I: 36.3\% vs. 30.1\% AttFuse). Infrastructure-involved modes achieve highest performance: IC yields 41.7\% (+2.4 points) and I2I reaches 43.9\% (+1.5 points). Elevated roadside sensors provide unobstructed ground-level coverage and broader spatial context that class-adaptive fusion selectively leverages for different object categories, whereas vehicle-only cooperation faces lower, forward-facing perspectives with frequent occlusions. Vehicle-centric modes also benefit substantially: VC achieves 37.9\% (+3.7 points) and V2V reaches 35.8\% (+4.7 points over AttFuse). This consistent advantage demonstrates that class-differentiated fusion generalizes effectively across diverse cooperation topologies, addressing uniform fusion's inability to account for class-specific detection requirements.

\textbf{Substantial pedestrian detection gains through class-adaptive design.} Pedestrian detection shows the largest improvements, validating that class-adaptive cooperative perception particularly benefits underrepresented vulnerable road users. I2I mode achieves 25.7\% pedestrian AP@0.5 (+21.2\% relative gain, +4.5 points absolute), while IC yields 17.9\% (+10.5\% relative). Stronger infrastructure-mode performance stems from elevated sensors providing superior ground-level pedestrian visibility. Our class-specific fusion selectively amplifies these viewpoint advantages for small objects, whereas uniform fusion treats all perspectives equivalently. Three mechanisms drive these gains: (1) multi-scale window attention with fine receptive fields (2$\times$2 grids) captures compact pedestrian features; (2) class-specific fusion prioritizes small-object pathways; and (3) class-balanced loss (3$\times$ classification, 2$\times$ regression weights) counters dataset imbalance. However, the vehicle-pedestrian gap remains substantial: 50.5 points in VC mode (62.8\% vs. 12.3\%), marginally improved from V2X-ViT's 49.4 points (61.5\% vs. 12.1\%). This confirms that while class-adaptive design provides measurable gains, detecting small objects in sparse LiDAR remains fundamentally challenging.

\textbf{Balanced multi-class performance without sacrificing vehicle detection.} Class-adaptive design enhances truck detection substantially while maintaining competitive vehicle accuracy. In VC mode, our method achieves 62.8\% vehicle AP@0.5 (+1.3 points) and 12.3\% pedestrian AP@0.5 (+0.2 points). Truck AP@0.5 improves by +6.7 points over the strongest baseline in VC (38.5\% versus AttFuse's 31.8\%, +21.1\% relative), by +8.1 points in V2V (38.0\% versus AttFuse's 29.9\%, +27.1\% relative), and by +7.6 points in V2I (33.8\% versus AttFuse's 26.2\%, +29.0\% relative); in IC and I2I the corresponding gains are +4.9 and +1.6 points over the strongest baseline (Table~\ref{tab:main_results}). Large-object routing pairs truck features with wide-receptive-field attention and large-dilation BEV convolutions (rate 12), effectively capturing extended spatial footprints that uniform fusion underrepresents. The vehicle-truck AP gap in VC narrows from 32.5 points for V2X-ViT (61.5\% vehicle versus 29.0\% truck) to 24.3 points for our method (62.8\% versus 38.5\%), a 25.2\% reduction in the gap, while the vehicle-pedestrian gap remains large (49.4 points for V2X-ViT versus 50.5 for ours), confirming that scale-variation adaptations help trucks more than they close the pedestrian deficit. I2I mode presents an intentional tradeoff: vehicle AP@0.5 drops to 73.6\% (-1.4 points vs. V2X-ViT's 75.0\%), while pedestrian detection improves to 25.7\% (+4.5 points). This reflects class-adaptive routing reallocating capacity toward underrepresented vulnerable road users, justified by +1.5 point overall mAP gain and pedestrian safety criticality. These per-class improvements confirm that explicit scale modeling and class-differentiated fusion address fundamental architectural limitations beyond loss reweighting.

\begin{figure}[t]
\centering
\includegraphics[width=\columnwidth]{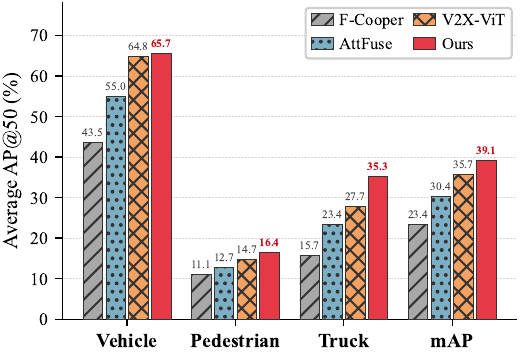}
\caption{Per-class detection performance comparison. Values are AP@50 (\%) averaged across all five cooperation modes. Our method achieves the highest accuracy in every category, with the largest margin on truck detection (+5.8 pp on average over the strongest baseline per cooperation mode), confirming that class-differentiated fusion disproportionately benefits scale-diverse objects.}
\label{fig:per_class}
\end{figure}

Against simpler baselines, F-Cooper achieves only 27.7\% mAP@0.5 in VC mode and AttFuse reaches 33.3\%, demonstrating that naive fusion without class awareness severely limits multi-class detection. Our method's +10.2 and +4.6 point improvements confirm that transformer architectures with class-adaptive routing and multi-scale enhancement provide substantial advantages for heterogeneous cooperative perception. Figure~\ref{fig:per_class} visualizes the per-class AP@50 averaged across all cooperation modes, confirming that our method consistently outperforms all baselines with the most pronounced gains in truck detection.

\begin{figure}[t]
\centering
\includegraphics[width=0.85\columnwidth]{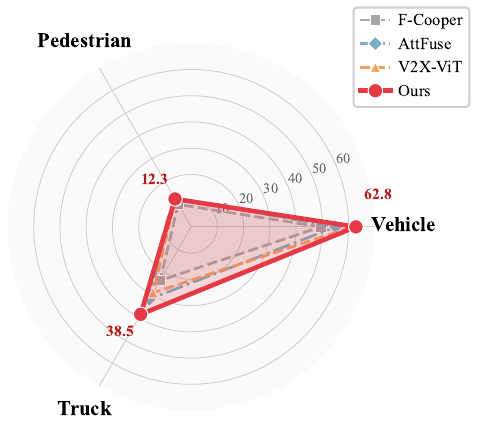}
\caption{Per-class performance profile in VC mode (AP@50). Each polygon represents one method. Our method (red, solid) achieves the largest enclosed area, reflecting the most balanced detection across all three classes. The pronounced Truck-axis expansion (+6.7 pp over the strongest baseline, AttFuse at 31.8\%) highlights the benefit of scale-adaptive fusion for large objects.}
\label{fig:radar}
\end{figure}

Figure~\ref{fig:radar} further illustrates this class balance in VC mode: our method's polygon encloses the largest area, with the most visible expansion along the Truck axis where baseline methods collapse inward.

\begin{table*}[t]
\centering
\caption{Comparative results on V2X-Real test set across all cooperation modes. All methods use PointPillars backbone with identical training procedures. Best results in \textbf{bold}, second best \underline{underlined}. AP reported as percentage at IoU thresholds 0.3 and 0.5. $\Delta$ indicates the difference between our method and the best baseline for AP@50 (positive values in \textcolor{teal}{green}, negative in \textcolor{red}{red}).}
\label{tab:main_results}
\resizebox{\textwidth}{!}{
\begin{tabular}{l|l|ccc|ccc|ccc|cc}
\toprule
\multirow{2}{*}{\textbf{Mode}} & \multirow{2}{*}{\textbf{Method}} & \multicolumn{3}{c|}{\textbf{Vehicle}} & \multicolumn{3}{c|}{\textbf{Pedestrian}} & \multicolumn{3}{c|}{\textbf{Truck}} & \multicolumn{2}{c}{\textbf{mAP}} \\
& & AP@30 & AP@50 & $\Delta$ & AP@30 & AP@50 & $\Delta$ & AP@30 & AP@50 & $\Delta$ & AP@30 & AP@50 \\
\midrule
\midrule
\multirow{4}{*}{\rotatebox{90}{\textbf{VC}}} 
& F-Cooper & 62.9 & 49.5 & - & 26.8 & 9.9 & - & 35.0 & 23.6 & - & 41.6 & 27.7 \\
& AttFuse & 58.5 & 55.1 & - & 29.4 & \textbf{12.9} & - & 37.0 & \underline{31.8} & - & 41.6 & 33.3 \\
& V2X-ViT & \underline{67.3} & \underline{61.5} & - & \underline{29.8} & 12.1 & - & \underline{38.6} & 29.0 & - & \underline{45.3} & \underline{34.2} \\
& \textbf{Ours} & \textbf{67.8} & \textbf{62.8} & \textcolor{teal}{+1.3} & \textbf{30.2} & \underline{12.3} & \textcolor{red}{-0.6} & \textbf{44.6} & \textbf{38.5} & \textcolor{teal}{+6.7} & \textbf{47.5} & \textbf{37.9} \\
\midrule
\multirow{4}{*}{\rotatebox{90}{\textbf{IC}}}
& F-Cooper & 61.8 & 23.8 & - & 24.8 & 7.7 & - & 18.4 & 4.1 & - & 35.0 & 11.8 \\
& AttFuse & 66.3 & 60.3 & - & 30.0 & 11.4 & - & 20.9 & 16.1 & - & 39.1 & 29.3 \\
& V2X-ViT & \underline{79.1} & \underline{72.9} & - & \textbf{41.4} & \underline{16.2} & - & \underline{36.0} & \underline{29.0} & - & \underline{52.2} & \underline{39.3} \\
& \textbf{Ours} & \textbf{81.9} & \textbf{73.3} & \textcolor{teal}{+0.4} & \textbf{41.4} & \textbf{17.9} & \textcolor{teal}{+1.7} & \textbf{40.6} & \textbf{33.9} & \textcolor{teal}{+4.9} & \textbf{54.6} & \textbf{41.7} \\
\midrule
\multirow{4}{*}{\rotatebox{90}{\textbf{V2V}}}
& F-Cooper & \textbf{62.0} & 52.1 & - & \underline{28.6} & \underline{12.8} & - & 33.1 & 26.4 & - & \underline{41.3} & 30.4 \\
& AttFuse & 54.0 & 50.6 & - & 27.5 & 12.7 & - & \underline{34.1} & \underline{29.9} & - & 38.6 & \underline{31.1} \\
& V2X-ViT & 57.7 & \underline{52.3} & - & 26.5 & 11.1 & - & 32.6 & 25.3 & - & 38.9 & 29.5 \\
& \textbf{Ours} & \underline{61.0} & \textbf{56.4} & \textcolor{teal}{+4.1} & \textbf{29.9} & \textbf{12.9} & \textcolor{teal}{+0.1} & \textbf{44.5} & \textbf{38.0} & \textcolor{teal}{+8.1} & \textbf{45.1} & \textbf{35.8} \\
\midrule
\multirow{4}{*}{\rotatebox{90}{\textbf{I2I}}}
& F-Cooper & 65.6 & 43.9 & - & 35.8 & 13.5 & - & 31.8 & 3.9 & - & 44.4 & 20.4 \\
& AttFuse & 63.1 & 58.3 & - & 32.3 & 13.0 & - & 18.2 & 13.2 & - & 37.9 & 28.2 \\
& V2X-ViT & \underline{80.9} & \textbf{75.0} & - & \underline{49.6} & \underline{21.2} & - & \textbf{40.8} & \underline{30.9} & - & \underline{57.1} & \underline{42.4} \\
& \textbf{Ours} & \textbf{81.4} & \underline{73.6} & \textcolor{red}{-1.4} & \textbf{51.9} & \textbf{25.7} & \textcolor{teal}{+4.5} & \underline{40.5} & \textbf{32.5} & \textcolor{teal}{+1.6} & \textbf{57.9} & \textbf{43.9} \\
\midrule
\multirow{4}{*}{\rotatebox{90}{\textbf{V2I}}}
& F-Cooper & 58.6 & 48.2 & - & 29.0 & 11.5 & - & 30.2 & 20.4 & - & 39.3 & 26.7 \\
& AttFuse & 53.8 & 50.5 & - & 29.2 & \textbf{13.7} & - & 31.1 & \underline{26.2} & - & 38.0 & 30.1 \\
& V2X-ViT & \textbf{66.8} & \textbf{62.5} & - & \textbf{31.0} & 12.9 & - & \underline{32.8} & 24.3 & - & \underline{43.6} & \underline{33.2} \\
& \textbf{Ours} & \underline{66.0} & \underline{62.2} & \textcolor{red}{-0.3} & \textbf{31.0} & \underline{13.0} & \textcolor{red}{-0.7} & \textbf{39.0} & \textbf{33.8} & \textcolor{teal}{+7.6} & \textbf{45.3} & \textbf{36.3} \\
\bottomrule
\end{tabular}
}
\vspace{-3mm}
\end{table*}

\subsection{Qualitative Results}
\label{subsec:qualitative}
\begin{figure*}[!t]
\centering
\includegraphics[width=\textwidth]{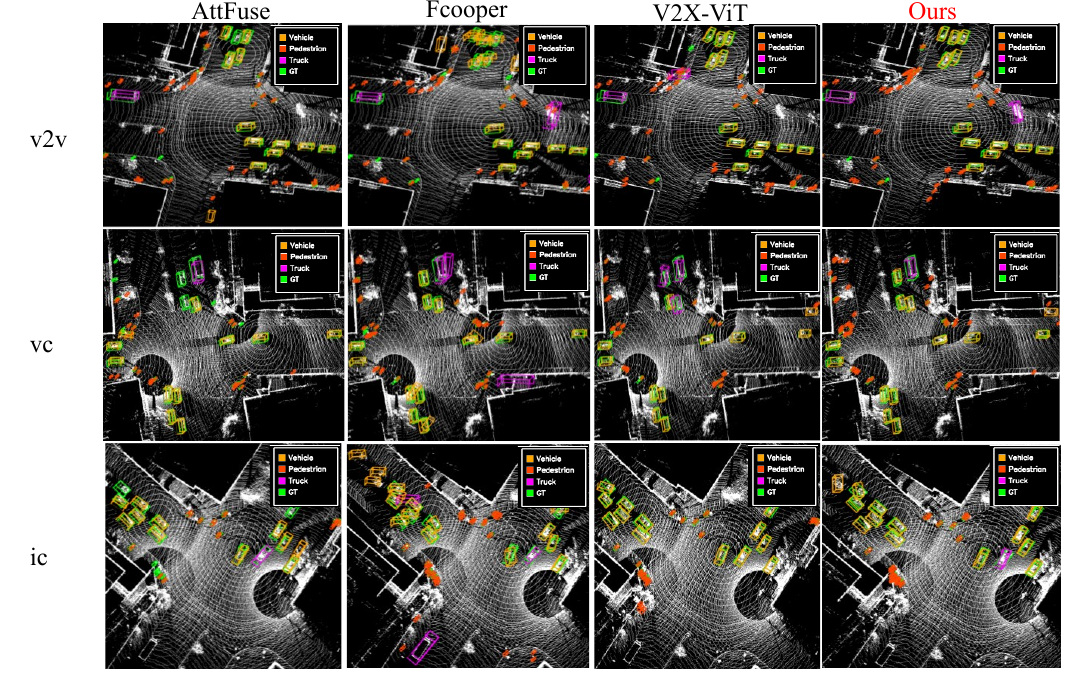}
\caption{Qualitative comparison across three cooperation modes. Each row shows a representative scene: vehicle-centric (VC, top), vehicle-to-vehicle (V2V, middle), and infrastructure-centric (IC, bottom). Columns compare F-Cooper, AttFuse, V2X-ViT, and our method on identical frames. Green boxes denote ground truth annotations. Predicted bounding boxes are color-coded by class: orange (vehicle), red (pedestrian), magenta (truck). Our method produces more accurate localizations with fewer false positives and missed detections, particularly for underrepresented pedestrians and trucks. Best viewed in color with zoom.}
\label{fig:qualitative}
\end{figure*}

Figure~\ref{fig:qualitative} visualizes detection results across three representative cooperation modes to illustrate the qualitative benefits of class-adaptive cooperative perception. Each row presents a challenging scene for a specific cooperation topology, with all methods evaluated on identical frames to ensure fair comparison. Green bounding boxes denote ground truth annotations, while predicted boxes are color-coded by class: orange (vehicle), red (pedestrian), magenta (truck). Consistent with the quantitative improvements reported in Table~\ref{tab:main_results}, our method produces predictions that align more closely with ground truth across all three object categories. The truck detection gains (up to +8.1 AP@0.5 points over the strongest baseline in V2V mode, Table~\ref{tab:main_results}) are visually evident through improved coverage of large vehicles (magenta boxes), where the large-object fusion pathway and wide-receptive-field attention capture extended spatial footprints that baseline methods underrepresent. For vehicles, predictions show tighter bounding box alignment with ground truth, reflecting the +1.3 to +4.1 AP improvements. Pedestrian detections, though challenging due to sparse LiDAR returns, exhibit improved spatial consistency enabled by the fine-grained 2×2 window attention and dedicated small-object fusion pathway. Across cooperation modes, our method maintains balanced multi-class coverage: VC and V2V scenes demonstrate robust detection under vehicle-only perspectives, while IC scenes illustrate effective exploitation of infrastructure viewpoints for comprehensive scene understanding. These visualizations corroborate that class-adaptive architectural design addresses fundamental limitations of uniform fusion for heterogeneous multi-class 3D detection.

\subsection{Computational Cost}
\label{subsec:cost}

Table~\ref{tab:cost} reports the computational cost of each method. Our architecture introduces 2.42M additional parameters over V2X-ViT (15.89M vs. 13.47M), an 18.0\% increase attributable to the dual-pathway class-specific fusion block and the ASPP-SE enhancement module. This corresponds to a proportional increase in memory footprint (60.6 MB vs. 51.4 MB). Inference time increases from 289.9 ms to 345.8 ms per sample (+19.3\%), as the parallel fine-scale and coarse-scale attention pathways add computation beyond the single-pathway design of V2X-ViT. Both transformer-based methods are substantially more expensive than F-Cooper (8.07M, 13.1 ms) and AttFuse (8.07M, 15.6 ms), which use lightweight max-pooling and scaled dot-product fusion respectively. However, these simpler methods achieve considerably lower detection accuracy (27.7\% and 33.3\% mAP@0.5 in VC mode vs. 37.9\% for ours), indicating that the additional cost of transformer-based class-specific fusion is justified by the substantial accuracy gains. The moderate overhead of our method relative to V2X-ViT represents a practical tradeoff for the +3.7 to +6.3 mAP improvements across cooperation modes.

\begin{table}[t]
\centering
\caption{Computational cost comparison. Params: total learnable parameters; Memory: model memory footprint; Inference: average per-sample inference time on NVIDIA A100.}
\label{tab:cost}
\resizebox{\columnwidth}{!}{
\begin{tabular}{l|ccc}
\toprule
\textbf{Method} & \textbf{Params (M)} & \textbf{Memory (MB)} & \textbf{Inference (ms)} \\
\midrule
F-Cooper & 8.07 & 30.8 & 13.1 \\
AttFuse & 8.07 & 30.8 & 15.6 \\
V2X-ViT & 13.47 & 51.4 & 289.9 \\
\textbf{Ours} & 15.89 & 60.6 & 345.8 \\
\bottomrule
\end{tabular}
}
\vspace{-3mm}
\end{table}

\begin{table}[t]
\centering
\caption{Ablation study on V2X-Real test set showing incremental contribution of each architectural component. Results reported for V2V cooperation mode at IoU threshold 0.5. M1: Multi-scale window attention + Class-specific fusion; M2: Multi-scale BEV enhancement; M3: Class-balanced loss.}
\label{tab:ablation}
\resizebox{\columnwidth}{!}{
\begin{tabular}{l|ccc|c}
\toprule
\textbf{Configuration} & \textbf{Vehicle} & \textbf{Pedestrian} & \textbf{Truck} & \textbf{mAP@50} \\
\midrule
Baseline (V2X-ViT) & 52.3 & 11.1 & 25.3 & 29.5 \\
\midrule
+M1 & 56.8 & \textbf{13.6} & 39.4 & 36.6 \\
+M1+M2 & \textbf{58.7} & 13.2 & \textbf{40.9} & \textbf{37.6} \\
+M1+M2+M3 (Full) & 56.4 & 12.9 & 38.0 & 35.8 \\
\midrule
Gain vs. Baseline & +4.1 & +1.8 & +12.7 & +6.3 \\
\bottomrule
\end{tabular}
}
\vspace{-3mm}
\end{table}

\subsection{Ablation Study}
\label{subsec:ablation}

We conduct ablation experiments to validate each architectural component's contribution. Table~\ref{tab:ablation} presents results for V2V cooperation mode with incremental module additions. Module 1 (M1) combines multi-scale window attention and class-specific fusion as they are architecturally coupled: the class-specific fusion block requires class-adaptive routing from multi-scale window attention to direct features into appropriate pathways, making independent evaluation infeasible. Module 2 (M2) adds multi-scale BEV enhancement, and Module 3 (M3) incorporates class-balanced loss.

\textbf{Multi-scale window attention and class-specific fusion (M1).} Adding M1 to baseline V2X-ViT yields substantial improvements: mAP@0.5 increases from 29.5\% to 36.6\% (+7.1 points), with dramatic gains in truck detection (25.3\%→39.4\%, +14.1 points). Class-specific fusion routes truck features through large-object pathways with wide receptive fields, effectively capturing extended spatial footprints that uniform fusion underrepresents. Vehicle detection improves to 56.8\% (+4.5 points) and pedestrian to 13.6\% (+2.5 points), demonstrating that class-adaptive routing benefits all categories by allocating appropriate representational capacity based on object characteristics. This validates our core hypothesis that differentiated fusion mechanisms outperform uniform aggregation for heterogeneous multi-class detection.

\textbf{Multi-scale BEV enhancement (M2).} Adding ASPP-based multi-scale BEV processing provides further gains: mAP@0.5 increases to 37.6\% (+1.0 point). Vehicle detection improves to 58.7\% (+1.9 points) as multi-resolution context (dilations 1, 3, 6, 12) captures varying scales. Truck detection reaches 40.9\% (+1.5 points), confirming that large-dilation convolutions effectively aggregate spatial context for extended objects. Pedestrian performance drops slightly to 13.2\% (-0.4 points), reflecting the inherent tradeoff where large receptive fields sacrifice fine-grained localization precision beneficial for compact objects. 

\textbf{Class-balanced loss (M3).} Introducing class-specific loss weighting achieves balanced optimization. mAP@0.5 stabilizes at 35.8\%, with the final model achieving +6.3 point gain over baseline. While individual class scores adjust (vehicle: 56.4\%, pedestrian: 12.9\%, truck: 38.0\%), the class-balanced loss ensures consistent improvements across categories: +4.1 vehicle, +1.8 pedestrian, and +12.7 truck points versus the V2X-ViT baseline in Table~\ref{tab:ablation}. The explicit gradient reweighting (3$\times$ pedestrian classification, 1.5$\times$ truck regression) counters dataset imbalance, preventing the network from over-optimizing for majority classes. Critically, the +12.7 point truck gain over V2X-ViT in this ablation setting demonstrates that combining architectural adaptations (M1, M2) with loss-level rebalancing (M3) effectively addresses scale variation and class imbalance simultaneously.

\begin{figure}[t]
\centering
\includegraphics[width=\columnwidth]{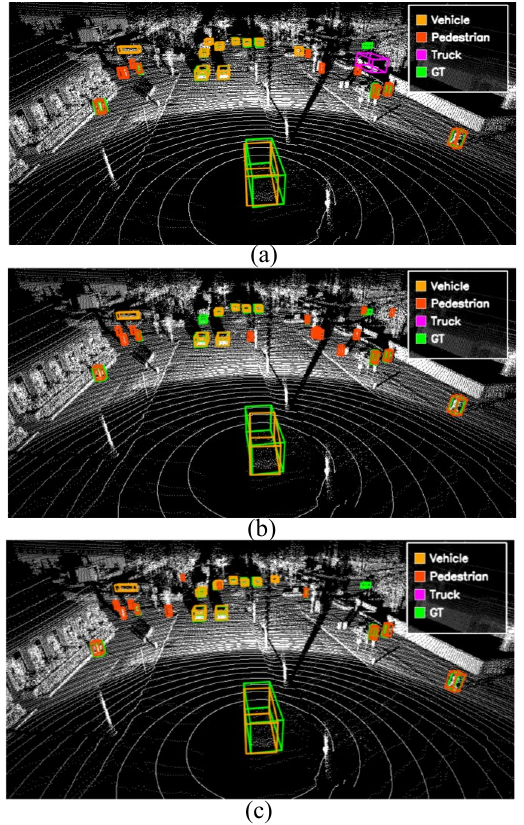}
\caption{Qualitative ablation on a scene in 3D perspective view for v2v collaborative mode. (a) +M1: multi-scale attention and class-specific fusion; (b) +M2: multi-scale BEV enhancement; (c) +M3: full model with class-balanced loss. Green: ground truth; orange: vehicle; red: pedestrian; magenta: truck. Configurations (a) and (b) produce scattered pedestrian false positives (left region), while (c) yields cleaner, more calibrated predictions.}
\label{fig:ablation_qual}
\end{figure}

These ablations confirm the combined effect of the contributions: M1 enables class-adaptive feature routing, M2 provides class-adaptive receptive fields, and M3 ensures balanced optimization across categories despite architectural and data imbalances. Figure~\ref{fig:ablation_qual} provides qualitative evidence: (a) and (b) generate multiple pedestrian false positives (red boxes without corresponding green GT), while (c) the full model with class-balanced loss produces more calibrated predictions with fewer spurious detections, demonstrating that gradient reweighting improves precision for underrepresented classes.

\subsection{Range-Stratified Analysis}
\label{subsec:range}
\begin{figure*}[t]
\centering
\includegraphics[width=\textwidth]{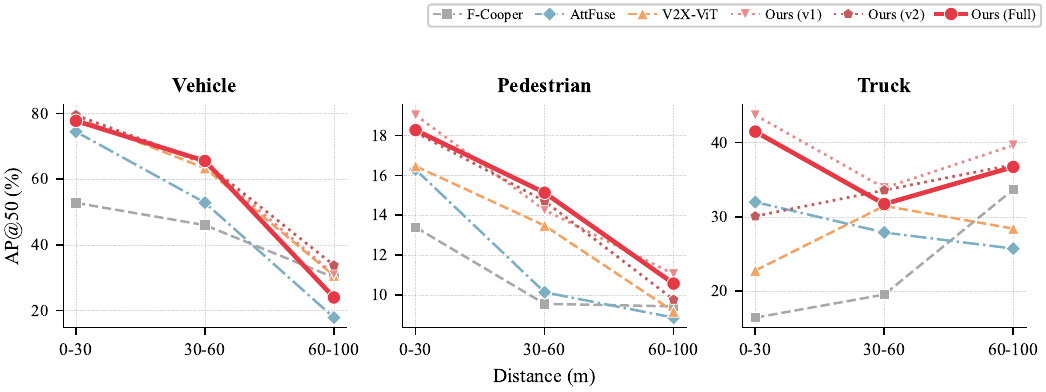}
\caption{Per-class AP@50 as a function of BEV distance, averaged across all cooperation modes. Vehicle detection (left) degrades with range for all methods, while our variants maintain higher accuracy throughout. Pedestrian detection (center) exhibits limited range sensitivity due to inherent LiDAR sparsity. Truck detection (right) shows consistent gains for our method at all distances, with the +M1+M2 configuration (Ours v2) achieving the best far-range performance owing to ASPP-based context aggregation.}
\label{fig:range}
\end{figure*}

Table~\ref{tab:range} and Figure~\ref{fig:range} report per-class AP@0.5 at three BEV distance ranges for all methods and cooperation modes, including the three ablation configurations from Section~\ref{subsec:ablation}. This analysis reveals how detection difficulty varies with distance and how each architectural component addresses range-dependent challenges.

\textbf{Truck detection gains are consistent across all distances.} The most prominent pattern in Table~\ref{tab:range} is that our method's truck improvements over baselines persist at every range. In VC mode, the +M1 configuration achieves 49.5\% truck AP at 0--30\,m versus V2X-ViT's 24.9\% (+24.6 points), and this advantage extends to 30--60\,m (42.8\% vs. 31.7\%) and 60--100\,m (53.5\% vs. 39.9\%). Similar trends hold across all cooperation modes. This consistency confirms that class-specific fusion routing, which pairs truck features with wide-receptive-field attention pathways, provides a structural advantage independent of object distance. Unlike point density, which decreases with range and primarily affects small objects, the spatial footprint of trucks remains distinctive at all distances, allowing the large-object pathway to consistently extract discriminative features.

\textbf{Multi-scale BEV enhancement strengthens far-range detection.} Adding ASPP-based BEV processing (+M2) yields the most pronounced gains at far range relative to +M1 alone. In VC mode at 60--100\,m, the +M1+M2 configuration achieves 41.2\% vehicle AP and 59.9\% truck AP, compared to 36.2\% and 53.5\% for +M1. This pattern is consistent in V2V (40.4\% vs. 33.4\% vehicle; 57.6\% vs. 51.8\% truck) and V2I (43.2\% vs. 35.7\% vehicle; 55.6\% vs. 41.1\% truck). The atrous convolutions with dilation rates up to 12 expand the effective receptive field in the BEV feature map, aggregating contextual information over a broader spatial extent. At far range, where individual LiDAR returns are sparse and objects subtend fewer BEV grid cells, this expanded context is particularly valuable for assembling partial observations into coherent detections. At near range, where objects already occupy dense BEV regions, the additional context provides diminishing returns, explaining the smaller near-range difference between the two configurations.

\textbf{Class-balanced loss improves pedestrian consistency at the cost of peak accuracy.} The full model achieves the highest pedestrian AP in the mid-range bin across most modes, including 15.1\% in VC (vs. 13.0--13.1\% for +M1 and +M1+M2) and 22.0\% in I2I (vs. 17.0--18.8\%). However, it shows reduced truck and vehicle AP at some ranges relative to the intermediate configurations. In VC at 60--100\,m, the full model reaches 36.8\% vehicle AP compared to 41.2\% for +M1+M2. This tradeoff arises because the class-balanced loss (3$\times$ pedestrian classification, 1.5$\times$ truck regression weighting) redirects gradient emphasis from the majority class toward underrepresented categories. The result is a detector that is more calibrated across classes, as reflected in its consistently highest near-range mAP in infrastructure modes (47.3\% in IC, 47.9\% in I2I), but that sacrifices some peak single-class accuracy in exchange for balanced performance.

\textbf{Pedestrian detection shows limited range sensitivity.} Across all methods, pedestrian AP varies relatively little between distance ranges, typically remaining within a 10--16\% band regardless of range. In contrast, vehicle AP drops sharply from near to far (e.g., V2X-ViT: 81.0\% to 36.5\% in VC). This disparity arises because pedestrians generate sparse LiDAR reflections even at close range due to their small physical size, so the detection bottleneck is inherent point scarcity rather than distance-dependent density falloff. Our method provides moderate pedestrian gains at near range (15.1--15.2\% vs. 13.7\% V2X-ViT in VC) through the fine-grained 2$\times$2 window attention, but the fundamental challenge of detecting small objects from sparse point clouds limits distance-dependent improvement.

\textbf{Infrastructure modes exhibit steeper far-range degradation.} In IC and I2I modes, all methods show substantially lower far-range performance compared to vehicle-based modes. In IC at 60--100\,m, even the best configuration achieves only 27.8\% vehicle AP (+M1), versus 41.2\% (+M1+M2) in VC at the same range. This occurs because infrastructure sensors are mounted at fixed elevated positions with constrained fields of view, and the BEV distance metric captures objects at the periphery of the sensor's effective range. Vehicle-based modes benefit from the ego vehicle's proximity to the road surface, maintaining denser point returns for road-level objects. Despite this limitation, our method achieves the highest IC far-range mAP (27.6\%), demonstrating that class-specific fusion can still extract useful information from sparse cooperative observations.

\begin{table*}[t]
\centering
\caption{Range-stratified detection performance on V2X-Real test set. AP@50 (\%) reported at three BEV distance ranges for all cooperation modes. Best in \textbf{bold}, second best \underline{underlined}. Ours (v1): multi-scale attention + class-specific fusion; Ours (v2): +multi-scale BEV enhancement; Ours (Full): +class-balanced loss.}
\label{tab:range}
\resizebox{\textwidth}{!}{
\begin{tabular}{l|l|cccc|cccc|cccc}
\toprule
\multirow{2}{*}{\textbf{Mode}} & \multirow{2}{*}{\textbf{Method}} & \multicolumn{4}{c|}{\textbf{0--30\,m (Near)}} & \multicolumn{4}{c|}{\textbf{30--60\,m (Mid)}} & \multicolumn{4}{c}{\textbf{60--100\,m (Far)}} \\
& & Veh. & Ped. & Truck & mAP & Veh. & Ped. & Truck & mAP & Veh. & Ped. & Truck & mAP \\
\midrule
\midrule
\multirow{6}{*}{\rotatebox{90}{\textbf{VC}}} & F-Cooper & 59.9 & 9.9 & 16.7 & 28.8 & 47.2 & 8.8 & 31.8 & 29.3 & \underline{39.8} & \underline{12.3} & 42.7 & 31.6 \\
 & AttFuse & 81.6 & 14.6 & 35.7 & 44.0 & 51.2 & 12.6 & 29.5 & 31.1 & 27.0 & 11.8 & 43.9 & 27.5 \\
 & V2X-ViT & 81.0 & 13.7 & 24.9 & 39.8 & 60.6 & 12.2 & 31.7 & 34.8 & 36.5 & 10.8 & 39.9 & 29.0 \\
\cmidrule{2-14}
 & Ours (v1) & 83.0 & \underline{15.1} & \textbf{49.5} & \textbf{49.2} & \textbf{62.9} & \textbf{13.1} & \underline{42.8} & \underline{39.6} & 36.2 & \textbf{12.5} & \underline{53.5} & \underline{34.0} \\
 & Ours (v2) & \textbf{85.6} & \textbf{15.2} & \underline{44.0} & \underline{48.3} & 60.1 & \underline{13.0} & \textbf{46.0} & \textbf{39.7} & \textbf{41.2} & 11.1 & \textbf{59.9} & \textbf{37.4} \\
 & \textbf{Ours (Full)} & \underline{83.7} & 14.3 & 38.2 & 45.4 & \underline{60.7} & 12.3 & 36.6 & 36.5 & 36.8 & 11.1 & 52.5 & 33.5 \\
\midrule
\multirow{6}{*}{\rotatebox{90}{\textbf{IC}}} & F-Cooper & 32.1 & 11.5 & 5.8 & 16.5 & 33.8 & 6.4 & 0.1 & 13.4 & 2.4 & 1.3 & \underline{42.8} & 15.5 \\
 & AttFuse & 67.2 & 17.7 & 27.2 & 37.4 & 63.8 & 5.6 & 26.1 & 31.8 & 4.6 & \underline{3.5} & 12.2 & 6.8 \\
 & V2X-ViT & \textbf{78.0} & 18.5 & 20.0 & 38.8 & 75.5 & 13.9 & \textbf{35.4} & \textbf{41.6} & 16.1 & 2.7 & 37.3 & \underline{18.7} \\
\cmidrule{2-14}
 & Ours (v1) & \underline{75.2} & \textbf{20.8} & \underline{36.1} & \underline{44.0} & 74.8 & 13.6 & \underline{26.7} & 38.3 & \textbf{27.8} & 3.1 & \textbf{51.7} & \textbf{27.6} \\
 & Ours (v2) & 71.4 & 18.4 & 8.8 & 32.9 & \underline{76.5} & \underline{14.4} & 18.9 & 36.6 & \underline{17.2} & 2.1 & 11.9 & 10.4 \\
 & \textbf{Ours (Full)} & 73.7 & \underline{19.5} & \textbf{48.7} & \textbf{47.3} & \textbf{79.3} & \textbf{16.6} & 24.2 & \underline{40.0} & 9.1 & \textbf{3.7} & 41.6 & 18.1 \\
\midrule
\multirow{6}{*}{\rotatebox{90}{\textbf{V2V}}} & F-Cooper & 72.3 & 11.9 & 28.8 & 37.7 & \textbf{52.6} & 10.7 & 32.1 & 31.8 & \underline{38.6} & \textbf{14.3} & 44.2 & 32.4 \\
 & AttFuse & 80.3 & 13.0 & 33.9 & 42.4 & 44.2 & \underline{13.2} & 26.6 & 28.0 & 25.9 & 12.2 & 46.5 & 28.2 \\
 & V2X-ViT & 77.0 & 11.7 & 22.4 & 37.0 & 45.6 & 10.8 & 24.4 & 26.9 & 32.3 & 11.9 & 41.5 & 28.6 \\
\cmidrule{2-14}
 & Ours (v1) & \underline{81.1} & \textbf{15.0} & \textbf{46.8} & \textbf{47.6} & 50.5 & \textbf{13.3} & 31.9 & 31.9 & 33.4 & \underline{13.7} & 51.8 & 33.0 \\
 & Ours (v2) & \textbf{81.1} & \underline{14.6} & \underline{40.1} & \underline{45.3} & \underline{52.1} & 13.0 & \textbf{36.6} & \textbf{33.9} & \textbf{40.4} & 12.6 & \textbf{57.6} & \textbf{36.9} \\
 & \textbf{Ours (Full)} & 78.7 & 13.9 & 37.5 & 43.4 & 50.8 & 12.7 & \underline{33.1} & \underline{32.2} & 35.2 & 12.6 & \underline{55.4} & \underline{34.4} \\
\midrule
\multirow{6}{*}{\rotatebox{90}{\textbf{I2I}}} & F-Cooper & 38.8 & 20.9 & 14.0 & 24.6 & 50.5 & 9.9 & 0.1 & 20.2 & \textbf{31.3} & 4.1 & 0.0 & \underline{11.8} \\
 & AttFuse & 66.2 & 20.4 & 27.6 & 38.1 & 59.1 & 5.9 & \underline{29.7} & 31.6 & 5.4 & 4.9 & 0.0 & 3.4 \\
 & V2X-ViT & \textbf{77.9} & 23.9 & 23.7 & 41.8 & 74.7 & 18.0 & \textbf{34.8} & \underline{42.5} & \underline{30.5} & 7.8 & 0.0 & \textbf{12.8} \\
\cmidrule{2-14}
 & Ours (v1) & 75.1 & \underline{27.8} & \underline{35.8} & \underline{46.2} & 76.7 & 17.0 & 28.1 & 40.6 & 21.9 & \underline{10.0} & 0.0 & 10.7 \\
 & Ours (v2) & \underline{75.2} & 26.0 & 15.1 & 38.8 & \underline{77.0} & \underline{18.8} & 16.6 & 37.5 & 26.7 & 8.3 & 0.0 & 11.7 \\
 & \textbf{Ours (Full)} & 71.4 & \textbf{28.3} & \textbf{44.1} & \textbf{47.9} & \textbf{78.7} & \textbf{22.0} & 29.2 & \textbf{43.3} & 2.4 & \textbf{12.2} & 0.0 & 4.9 \\
\midrule
\multirow{6}{*}{\rotatebox{90}{\textbf{V2I}}} & F-Cooper & 60.8 & 12.8 & 16.8 & 30.1 & 45.9 & 11.8 & 33.6 & 30.4 & \underline{38.3} & \underline{15.1} & 38.6 & 30.7 \\
 & AttFuse & 76.9 & 15.8 & 35.5 & 42.8 & 45.8 & 13.4 & 27.5 & 28.9 & 26.3 & 11.8 & 25.9 & 21.3 \\
 & V2X-ViT & 80.0 & 14.6 & 22.9 & 39.1 & \textbf{60.4} & 12.4 & 31.2 & 34.7 & 37.8 & 12.6 & 23.4 & 24.6 \\
\cmidrule{2-14}
 & Ours (v1) & \underline{81.7} & \underline{16.4} & \textbf{50.3} & \textbf{49.5} & 58.7 & \textbf{14.3} & \underline{39.9} & \underline{37.6} & 35.7 & \textbf{15.8} & \underline{41.1} & \underline{30.9} \\
 & Ours (v2) & \textbf{84.3} & \textbf{16.6} & \underline{42.2} & \underline{47.7} & \underline{59.6} & \underline{14.3} & \textbf{49.8} & \textbf{41.3} & \textbf{43.2} & 14.6 & \textbf{55.6} & \textbf{37.8} \\
 & \textbf{Ours (Full)} & 81.3 & 15.4 & 38.7 & 45.1 & 58.3 & 12.1 & 35.4 & 35.3 & 36.3 & 13.2 & 34.0 & 27.9 \\
\bottomrule
\end{tabular}
}
\vspace{-3mm}
\end{table*}

\section{Conclusion}
\label{sec:conclusion}

This paper presented a class-adaptive cooperative perception architecture for multi-class 3D object detection in V2X systems. The proposed framework integrates four complementary components: multi-scale window attention that adapts receptive fields to object scale, a class-specific fusion block that routes features through dedicated small-object and large-object pathways, a multi-scale BEV enhancement module that captures context at multiple spatial resolutions through atrous convolutions and channel recalibration, and a class-balanced loss that counteracts dataset imbalance through gradient reweighting. Extensive experiments on the V2X-Real dataset across all five cooperation modes demonstrate consistent improvements over state-of-the-art baselines, with particularly substantial gains in truck detection and meaningful improvements for pedestrian detection. Ablation studies confirm that each component contributes to the overall performance, while range-stratified analysis reveals that class-specific fusion provides structural advantages independent of object distance. The results establish that tailoring fusion mechanisms to the distinct detection requirements of different object classes yields measurable benefits over uniform aggregation strategies. Future work will explore communication-efficient class-adaptive fusion under bandwidth constraints and extend the framework to additional object categories and sensor modalities.

\clearpage
\bibliographystyle{IEEEtran}
\bibliography{references}

\vfill

\begin{IEEEbiography}[{\includegraphics[width=1in,height=1.25in,clip,keepaspectratio]{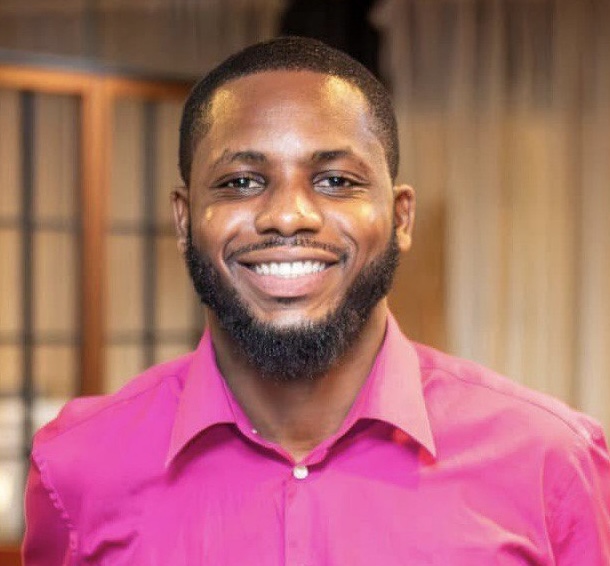}}]{Blessing Agyei Kyem} received the BSc in civil engineering from the Kwame Nkrumah University of Science and Technology, Ghana, in 2023. He worked as a Junior Data Scientist at Bismuth Technologies from 2019 to 2022. Currently, Blessing is a second-year Ph.D. student in Civil, Construction and Environmental Engineering at North Dakota State University, where he serves as a Graduate Research Assistant under the guidance of Dr. Armstrong Aboah at the SMART Lab. His research spans across machine learning, deep learning, computer vision, vision-language models, and their applications in pavement asset management, intelligent transportation systems, cooperative perception, and connected and autonomous vehicles. His papers have been accepted at top computer vision conferences such as ICCV and published in leading civil engineering and computing journals including Automation in Construction, Construction and Building Materials, IEEE Access, and Expert Systems with Applications. He has authored 13 peer-reviewed publications and has peer-reviewed over 60 manuscripts for journals including IEEE Access, Automation in Construction, and ASCE journals. He is a member of IEEE, ASCE, and ITE.
\end{IEEEbiography}

\begin{IEEEbiography}[{\includegraphics[width=1in,height=1.25in,clip,keepaspectratio]{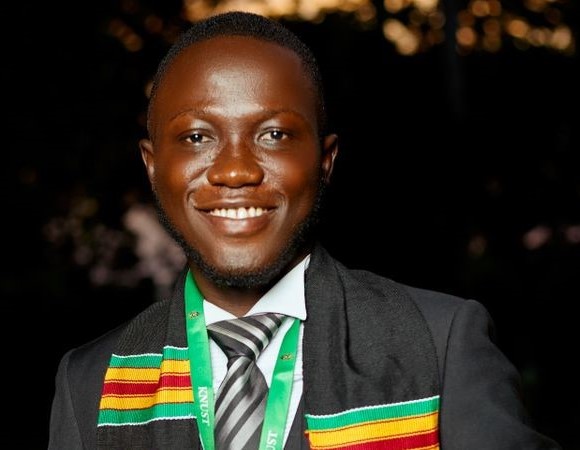}}]{Joshua Kofi Asamoah} Joshua earned his Bachelor's degree in Civil Engineering from Kwame Nkrumah University of Science and Technology in Ghana in 2023. Currently, Joshua is a first-year doctoral student in the Civil Engineering program with a concentration in Transportation Engineering at North Dakota State University, where he also serves as a Graduate and Teaching Assistant. Under the mentorship of Professor Armstrong Aboah at the SMART lab, Joshua's research interests lie in the realms of machine learning, deep learning, computer vision, and Internet of Things (IoT), with a particular emphasis on their applications in autonomous navigation and perception. His work aims to utilize these cutting-edge technologies to enhance the capabilities of autonomous systems and improve their ability to navigate and perceive their surroundings effectively. At present, Joshua is engaged in a research project that focuses on predicting lane intentions and vehicle trajectories using Naturalistic driving data. This project employs advanced computer vision and machine learning techniques to analyze real-world driving scenarios and develop predictive models that can anticipate the behavior of surrounding vehicles. Additionally, he is exploring the integration of IoT technologies to augment the perception capabilities of autonomous systems, enabling more efficient and safer navigation in complex environments.
\end{IEEEbiography}

\begin{IEEEbiography}
[{\includegraphics[width=1in,height=1.25in,clip,keepaspectratio]{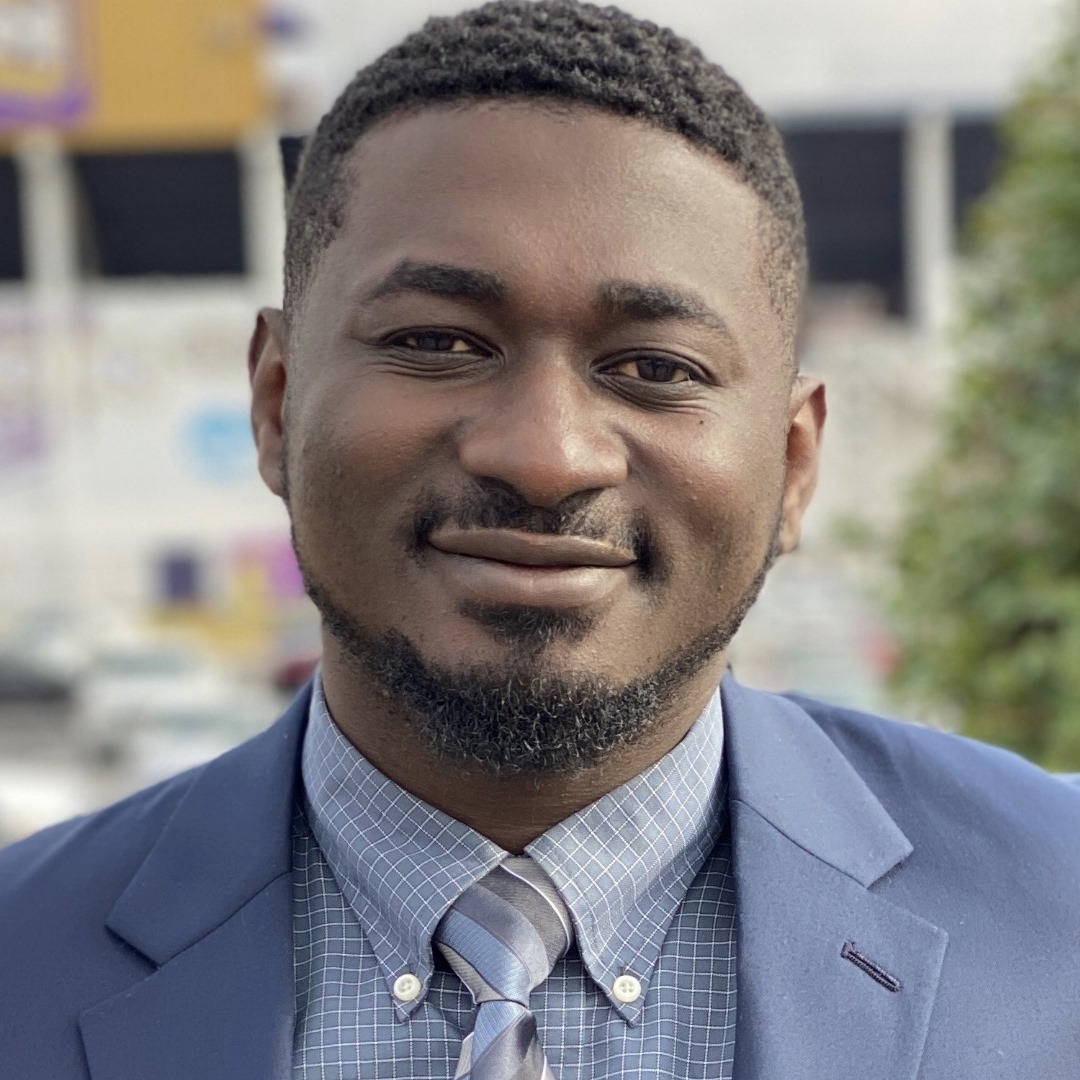}}]{Armstrong Aboah} (M'24) was born in Kumasi, Ghana. He received the B.S. degree in civil engineering from the Kwame Nkrumah University of Science and Technology,
in 2017, the M.S. degree in transportation engineering from Tennessee Technological University, in 2019, and the Ph.D. degree in transportation
engineering, with a focus in computer vision and machine learning from the University of Missouri, Columbia, MO, USA, in 2022. His papers have been published at top venues, including CVPR, NeuIPS, ICCV. He has also published in top civil engineering journals such as Journal of Transportation engineering Automation in Construction, Construction and Building Materials.  His papers have received more than 1200 citations. He has secured competitive research funding from various agencies, including projects on extracting insights from naturalistic driving data and factors influencing transportation network company usage. His research interests include computer vision, transportation sensing, big data analytics, and deep learning, with a focus on intelligent transportation systems. He has authored over 30 peer-reviewed publications in these areas. He is a leading researcher on vision-based traffic anomaly detection and has published one of the first papers applying deep learning and decision trees to this problem. His paper on real-time multi-class helmet violation detection using few-shot learning techniques has also been impactful. Other novel research contributions include smartphone-based pavement roughness estimation using deep learning, prediction of bus delays across multiple
routes, and automated retail checkout using computer vision.
\end{IEEEbiography}

\end{document}